\documentclass[10pt,twocolumn,letterpaper]{article}

\usepackage[pagenumbers]{cvpr} % arXiv / preprint with page numbers

\definecolor{cvprblue}{rgb}{0.21,0.49,0.74}
\usepackage[pagebackref,breaklinks,colorlinks,allcolors=cvprblue]{hyperref}

\usepackage{cuted}
\usepackage{booktabs,multirow,adjustbox}
\usepackage{graphicx}
\usepackage{caption}
\usepackage{subcaption}
\usepackage{array}
\usepackage{tabularx}
\usepackage{makecell}

\def\etal{\textit{et al}.}

\def\paperID{227}
\def\confName{3DV\xspace}
\def\confYear{2027\xspace}

\title{RF3DGS: Reflection-Free 3D Gaussian Splatting with Self-Priors}

\author{
Qiang Wen$^{1\dagger}$\quad
Guotao Meng$^{1\dagger}$\quad
Junhong Ruan$^{1\dagger}$\quad
Hongyu Liu$^{1}$\quad
Chenyang Lei$^{2}$\quad
Qifeng Chen$^{1}$\\
$^{1}$The Hong Kong University of Science and Technology\\
$^{2}$Huawei Research
}

\begin{document}
\maketitle

\begin{strip}
\centering
    \includegraphics[width=\textwidth]{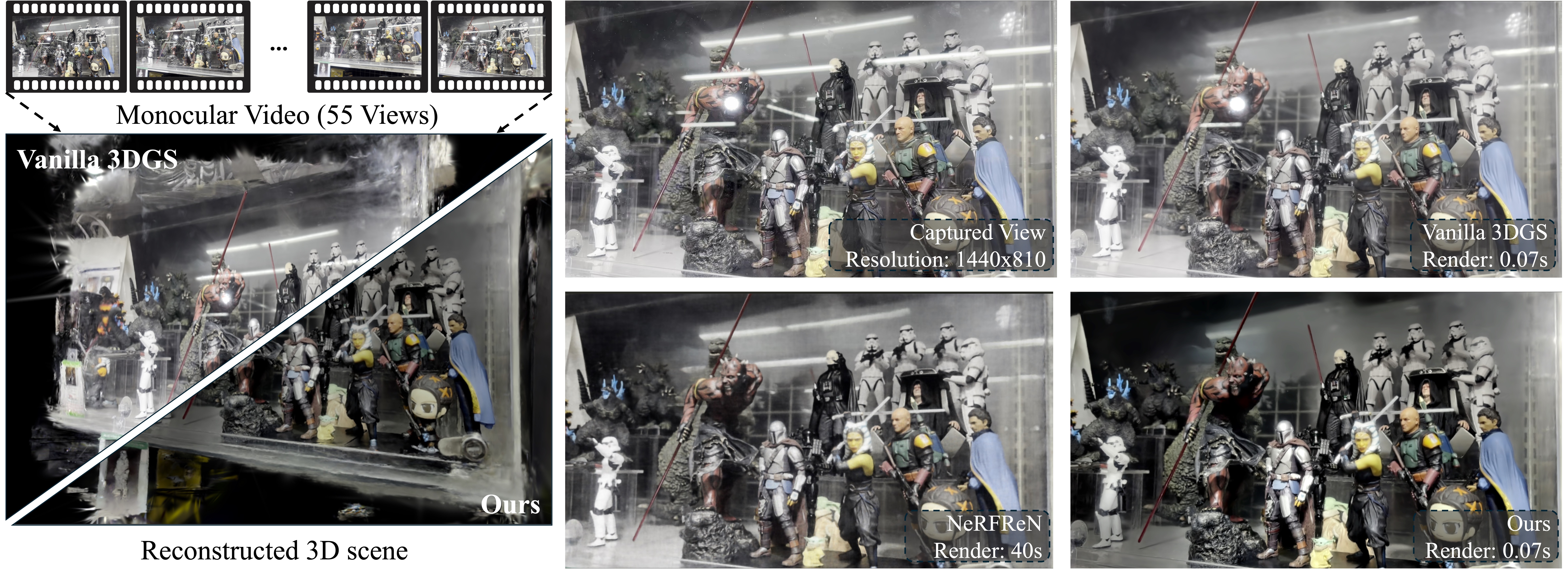}
    \captionof{figure}{Given a monocular video captured in a real-world scenario, the proposed method demonstrates its effective and efficient reflection removal performance. Compared with vanilla 3DGS~\cite{kerbl3Dgaussians}, the proposed method exhibits greater robustness in real-world scenarios. Compared with the state-of-the-art 3D reflection removal method, NeRFReN~\cite{Guo_2022_CVPR}, the proposed method not only proves effective but also excels in efficiency, enabling real-time rendering.}
    \label{fig: teaser}
\end{strip}

\begin{figure}[b]
\vspace{-1.5em}
\noindent\footnotesize\normalfont$^{\dagger}$ Joint first authors.
\end{figure}
\begin{abstract}
3D reflection removal is crucial in applications demanding high-fidelity visualization, such as museum exhibitions. In this work, we propose Reflection-Free 3D Gaussian Splatting (RF3DGS), a novel method for superior reflection removal. RF3DGS exclusively leverages a widely-adopted photometric prior for reflection removal: \textbf{\textit{that the minimum pixel intensity across multi-view images corresponding to the same 3D point strongly correlates with reflection-free information.}} However, generalizing this principle to a 3D Gaussian Splatting within a multi-view context is non-trivial. To address this, RF3DGS first optimizes a vanilla 3DGS model using a geometric prior to estimate a coarse, reflection-free depth map for each view. Leveraging these depth maps and photometric prior, a 3D warping process aggregates reflection-free information from auxiliary views to construct a clean transmission pseudo-label for a given reference view. This pseudo-label provides a robust self-supervised signal for optimization. We further introduce an Adaptive Twin Rendering (ATR) loss to synergistically utilize this pseudo-label while preserving high-frequency details from the original, reflection-degraded images. By adeptly fusing these internal 3D self-priors with the ATR loss, RF3DGS demonstrates state-of-the-art reflection removal performance across both synthetic and real-world scenarios. Furthermore, we establish the first 3D reflection removal benchmark, based on the Unity rendering engine, for rigorous quantitative evaluation.
\end{abstract}

\section{Introduction}
\label{sec:intro}
Capturing photos with undesirable reflections is common in daily scenarios, such as exhibitions shown in Fig.~\ref{fig: teaser}. Reflection removal seeks to restore clean transmission for better visual quality and improved downstream computer vision performance~\cite{redmon2016you}.

However, removing reflections in real-world scenarios remains ill-posed. Learning-based methods improve robustness with additional priors, e.g., edge maps~\cite{fan2017generic}, polarization~\cite{lei2020polarized}, textual prompts~\cite{zhong2024language, ldiffer_hong}, and flash/no-flash cues~\cite{lei2023robust}. Due to limited scene coverage in current datasets~\cite{zhang2018single, zhu2024revisiting, wan2017benchmarking, li2020single, lei2020polarized, niklaus2021learned, wang2022background, Liu_CVPR_2020}, such methods still suffer from domain gap and often fail under strong reflections or non-ideal guidance. Multi-frame methods~\cite{guo2014robust, Xue2015ObstructionFree, gai2011blind, li2013exploiting, alayrac2019visual, chugunov2024neural} can aggregate clean information across frames, yet priors from limited neighbors (e.g., optical flow) remain too weak for strong reflections or significant motions~\cite{li2013exploiting, chugunov2024neural}.

Thanks to progress in 3D neural scene representations~\cite{mildenhall2020nerf, kerbl3Dgaussians}, 3D reflection removal methods~\cite{Guo_2022_CVPR, flash_splat, wang2024dc, gao2024planar, zhu2022neural} show strong robustness via multi-view consistency and crafted assumptions. Both NeRF-based (e.g., TNeRF~\cite{zhu2022neural}, NeRFReN~\cite{Guo_2022_CVPR}) and 3DGS-based methods (e.g., DC-Gaussian~\cite{wang2024dc}, Flash-Splat~\cite{flash_splat}) perform well. However, NeRFReN’s long training and rendering times hinder practicality, while pixel-wise optimization limits global perceptual quality. Flash-Splat exploits real-time 3DGS rendering~\cite{kerbl3Dgaussians}, but its reliance on flash/no-flash cues~\cite{lei2023robust, xia2021deep, xia2021dark} in \textbf{RAW} images complicates capture and SfM; these cues also weaken as camera–scene distance grows.

To inherit the robustness and overcome the weaknesses of current 3D reflection removal methods, we propose a novel reflection-free 3DGS method for the 3D reflection removal task.
The proposed method simplifies the capture process by requiring only a monocular \textbf{sRGB} video, easily obtained with any mobile phone or DSLR camera, without relying on specialized photography techniques such as capturing flash or no-flash RAW images~\cite{flash_splat}.
Inspired by Li~\etal~\cite{li2013exploiting}, we leverage the \textbf{Photometric Prior} in real-world reflection scenarios: for a given 3D scene point, the minimum intensity value across all captured views strongly corresponds to reflection-free information. Since pixel values from the captured views must be accurately projected back into 3D space, obtaining precise depth maps is crucial to effectively apply the photometric prior.
Benefiting from explicit depth estimation in 3DGS, we can efficiently generate depth maps for each captured view by optimizing a vanilla 3DGS model under reflection conditions.
To ensure the depth maps contain as pure reflection-free depth information as possible, we incorporate a \textbf{Geometric Prior}—also used by NeRFReN~\cite{Guo_2022_CVPR}—within 3DGS, which encourages smoothness in the rendered depth maps.
This geometric prior assumes that the depth map of the transmitted component should be locally smoother than that of the reflection-degraded one.
Utilizing these refined depth maps, we project all captured views back into the 3D space and apply the \textit{photometric prior}, then re-render each view with preliminary clean transmission information.
Due to the flaws of depth maps and imperfect projections, current reflection-free views still have inevitable artifacts, as shown in Fig.~\ref{fig: 3D_warp}.
Directly optimizing the 3DGS on them leads to poor 3D consistency and color distortion.
To address this issue, the proposed method goes into the comprehensive optimization where original captured views and preliminary reflection-free views are jointly treated as supervision during 3DGS optimization.
We propose an \textbf{A}daptive \textbf{T}win \textbf{R}ending (ATR) loss to achieve this joint supervision, encouraging the 3DGS model to effectively capture the clean transmission information from preliminary reflection-free views and maintain the original 3D consistency for a reflection-degraded scene.

Furthermore, recognizing that existing benchmarks provide limited scene diversity—for example, TNeRF~\cite{zhu2022neural} includes only four scenes—for quantitatively evaluating 3D reflection removal methods, we introduce a comprehensive 3D reflection removal benchmark aimed at significantly enhancing evaluation diversity.
Since capturing videos with and without reflections along the exact same camera trajectory is challenging, we leverage the precise camera control capabilities of the modern rendering engine Unity~\cite{haas2014history} to construct our benchmark.
To guarantee the diversity of our benchmark, we build 3D scenes with different settings (background \& reflection scenes, glass thickness, transparency and reflection strength), resulting in 45 cases for evaluation.
Besides, we additionally collect real-world cases for qualitative evaluation.
Extensive experiments demonstrate that the proposed method significantly outperforms previous works on the proposed benchmark and across various real-world scenarios. This strong robustness and simplicity of our method indicate its potential for practical applications.

\begin{figure*}[!t]
\includegraphics[width=\textwidth]{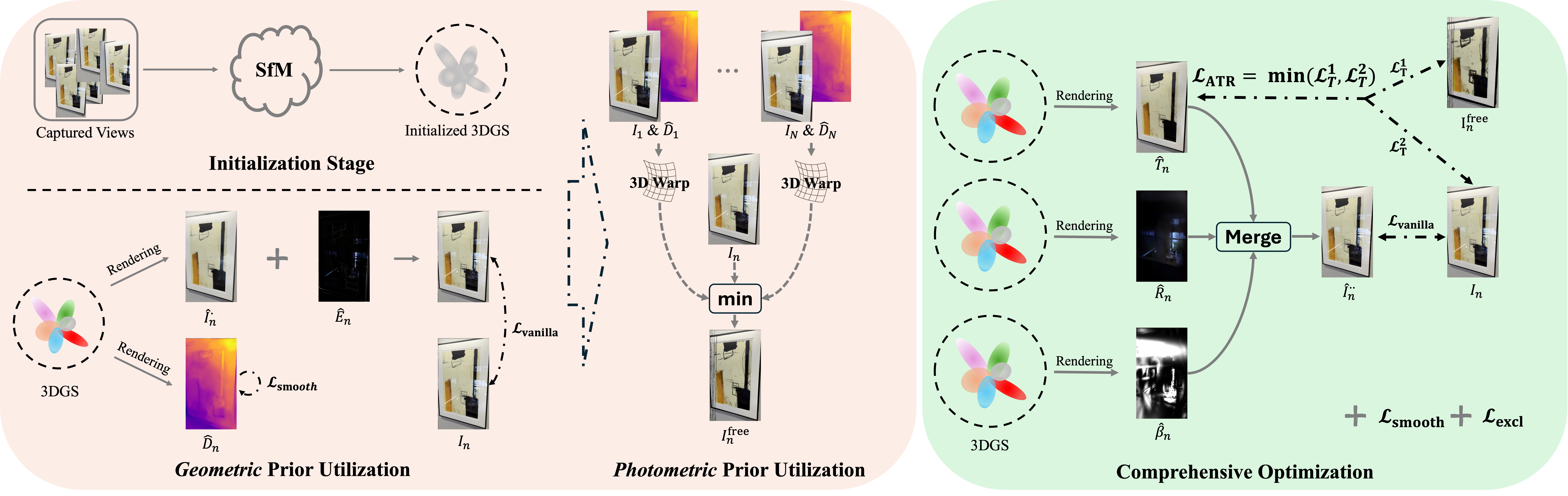}
\vspace{-5mm}
\caption{An overview of the proposed method. 1) During \textit{Initialization}, all captured views are processed through structure-from-motion (SfM) to generate an initial 3D scene point cloud, which is then used to initialize a 3DGS model; 2) During \textit{Geometric Prior Utilization}, the 3DGS model is optimized using learnable residual maps $\hat{E}_{n}$ combined with the depth smoothness geometric prior to generate coarse reflection-free depth maps $\hat{D}_{n}$; 3) During \textit{Photometric Prior Utilization}, using the reflection-free depth maps, all captured views are projected back onto the 3D scene point cloud. Subsequently, for re-rendering each specific view, a minimum value-based photometric prior is applied to extract preliminary clean transmission information; 4) During \textit{Comprehensive Optimization}, three 3DGS models are jointly optimized for producing transmission image $\hat{T}_{n}$, reflection image $\hat{R}_{n}$ and reflectance factor $\hat{\beta}_{n}$, respectively.}
\label{fig: framework}
\end{figure*}
\section{Related Work}
\label{sec:related}
\subsection{Single-Image Reflection Removal}
With the rapid development of deep learning techniques, learning-based methods~\cite{wang2023personalized, dong2021location, hu2021trash, fan2017generic, ldiffer_hong, zhong2024language, li2020single, hu2023single, ma2019learning, prasad2021v, wan2018crrn, wan2019corrn, wan2020reflection, wei2019single, wen2019single, yang2018seeing, zhang2018single, zheng2021single} are chosen as the first alternative compared with early traditional methods~\cite{arvanitopoulos2017single, guo2014robust, Xue2015ObstructionFree, levin2007user, li2014single, shih2015reflection, zheng2021single, yang2019fast}.
Fan~\etal~\cite{fan2017generic} propose the pioneer learning-based method, CEILNet, which first estimates an edge map as the assisted cue for the subsequent transmission layer estimation.
Zhang~\etal~\cite{zhang2018single} propose to utilize perceptual loss~\cite{johnson2016perceptual} to train a deep network by considering misalignments,  which are caused by glass refraction, between the estimated and ground-truth transmission layers.
Wen~\etal~\cite{wen2019single} propose the generative-based reflection synthesis and removal networks for training the removal network based on the nonlinear synthetic data.
Kim~\etal~\cite{kim2020single} propose a physically based rendering pipeline for data synthesis for subsequent removal network training.
Hu~\etal~\cite{hu2023single} propose to capture a learnable residue term during input image decomposition, which guides a further separation between transmission and reflection layers.
Zhu~\etal~\cite{zhu2024revisiting} propose a cascaded method, using their proposed maximum reflection filter to locate reflection for guiding subsequent removal.
Most recently, Zhong~\etal~\cite{zhong2024language} and Hong~\etal~\cite{ldiffer_hong} propose language-guided methods by utilizing text prompts to aid reflection removal.

While learning-based methods demonstrate strong performance in benchmark evaluations, the challenge of domain adaptation limits their effectiveness in real-world daily scenarios.

\subsection{Multi-frame Reflection Removal}
Compared to single-image methods, multi-frame methods~\cite{guo2014robust, Xue2015ObstructionFree, gai2011blind, li2013exploiting, alayrac2019visual, Liu_CVPR_2020} show their superiority with exploiting useful information from multiple frames.
Alayrac~\etal~\cite{alayrac2019visual} and Liu~\etal~\cite{Liu_CVPR_2020} propose learning-based methods which are trained on synthetic data.
Such methods still suffer from domain-gap problems when they are applied on real-world scenarios.
Despite learning-based methods, traditional multi-frame methods focus on exploiting image structural priors~\cite{guo2014robust, li2013exploiting} or motion priors~\cite{gai2011blind, Xue2015ObstructionFree} for separating transmission and reflection layers.
Most recently, Chugunov~\etal~\cite{chugunov2024neural} propose neural spline fields for burst frames reflection removal.
However, these multi-frame methods face limitations when dealing with significant camera motion, particularly with large changes in viewing angles.
Other works seek to enhance reflection removal performance by capturing additional data through optical priors, such as polarization-based methods~\cite{lei2020polarized}, flash/no-flash techniques~\cite{lei2023robust}, and RAW-domain approaches~\cite{reflection_raw_eric}.
These methods are limited by their dependence on additional data acquisition, which increases complexity and may not be practical in most everyday situations.

\subsection{3D Reflection Removal}
There are some existing NeRF-based methods~\cite{zhu2022neural, gao2024planar, Guo_2022_CVPR} for the 3D reflection removal task. TNeRF~\cite{zhu2022neural} introduces a pre-trained transmission encoder~\cite{wei2019single} and recurring edge constraints as
guidance into the NeRF backbone to improve its reflection removal performance.
Guo~\etal~\cite{Guo_2022_CVPR} propose a NeRF-based method, NeRFReN, by adopting a similar image formulation as Shina~\etal~\cite{sinha2012image}.
The effectiveness of NeRFReN can be attributed to three main aspects: the incorporation of a depth smoothness prior to within the transmission layers, the implementation of bidirectional depth consistency, and the observation that reflections exhibit simpler geometrical characteristics compared to transmission layers.
Gao~\etal~\cite{gao2024planar} uses a similar architecture to NeRFReN but models planar surfaces and casts reflected rays to capture reflections, relying on a pre-trained segmentation model~\cite{kirillov2023segment} to initialize window surfaces.

There are some existing 3DGS-based methods~\cite{wang2024dc, flash_splat, song2025reflections}.
However, DC-Gaussian~\cite{wang2024dc} removes reflections from multi-view transmission images degraded by a \textit{single-view} reflection image, while our work tackles the more complex case of multi-view transmission images degraded by \textit{multi-view} reflection images.
The most recent and similar work to ours is Flash-Splat~\cite{flash_splat}.
Flash-Splat utilizes flash/no-flash cues~\cite{lei2023robust, xia2021deep, xia2021dark} for optimizing 3DGS~\cite{kerbl3Dgaussians} with unpaird flash/no-flash views in \textbf{RAW} format.
However, the major limitation of flash-based methods is that the effectiveness of flash cues rapidly diminishes as the distance between the camera and the target scene increases.
In this work, the proposed method does not need any additional photography techniques but multi-view captures that can be simply captured in a video mode using the \textbf{sRGB} format.
This prevents the proposed method from being limited by any constraints.
\section{Method}
In this work, we define the 3D reflection removal task as follows: Given a set of multi-view captures $\mathcal{I}_{N}=\{I_{n}, n\in N\}$, where $N$ denotes the view numbers, our objective is to achieve 3D reflection removal by reconstructing transmission and reflection scenes separately.

We follow prior work~\cite{Guo_2022_CVPR, flash_splat} and model each reflection-degraded observation as:
\begin{equation}
I_{n} = T_{n} + \beta_{n}  \circ R_{n},
\label{eq: degradation model}
\end{equation}
where $n$ denotes view index, $T_{n}$ $R_{n}$, and $\beta_{n}$ denote transmission image, reflection image and reflective fraction map.

We reconstruct the scene with 3D Gaussian Splatting (3DGS)~\cite{kerbl3Dgaussians}.
Each Gaussian is parameterized by a mean \(x \in \mathbb{R}^{3}\), a covariance \(\Sigma\) factored as \(\Sigma = \boldsymbol{R}\boldsymbol{S}\boldsymbol{S}^{T}\boldsymbol{R}^{T}\), opacity \(\alpha\), and view-dependent color \(c\) encoded by spherical harmonics.
A pixel color is obtained by front-to-back \(\alpha\)-blending of the \(\mathcal{N}\) overlapping Gaussians:
\begin{equation}
C = \sum_{i \in \mathcal{N}} c_{i}\alpha_{i}\prod_{j=1}^{i-1}(1-\alpha_{j}).
\end{equation}
Vanilla 3DGS is optimized with a combination of \(\mathcal{L}_{1}\) and D-SSIM:
\begin{equation}
\mathcal{L}_{\text{vanilla}} = \lambda \mathcal{L}_{1}(\hat{I}, I) + (1-\lambda)\mathcal{L}_{\text{D-SSIM}}(\hat{I}, I),
\end{equation}
with \(\lambda=0.2\). We reuse this photometric term in later stages and refer readers to~\cite{kerbl3Dgaussians} for rasterization and densification details.

\subsection{Reflection-Free 3D Gaussian Splatting}
In this work, we propose a novel reflection-free 3DGS method.
Unlike the recent 3DGS-based method, Flash-Splat~\cite{flash_splat}, which requires flash / no-flash multi-view RAW image captures, the proposed method needs no additional photography techniques but only multi-view sRGB images captured by a monocular video.
This allows the proposed method to be free from the effectiveness declining of flash / no-flash cues with distance increasing as mentioned in works~\cite{flash_splat, lei2023robust, xia2021deep, xia2021dark}.

\subsubsection{Geometric Prior Utilization}
\begin{figure}[h]
  \centering
  \begin{minipage}{0.45\textwidth}
    \centering
    \includegraphics[width=\textwidth]{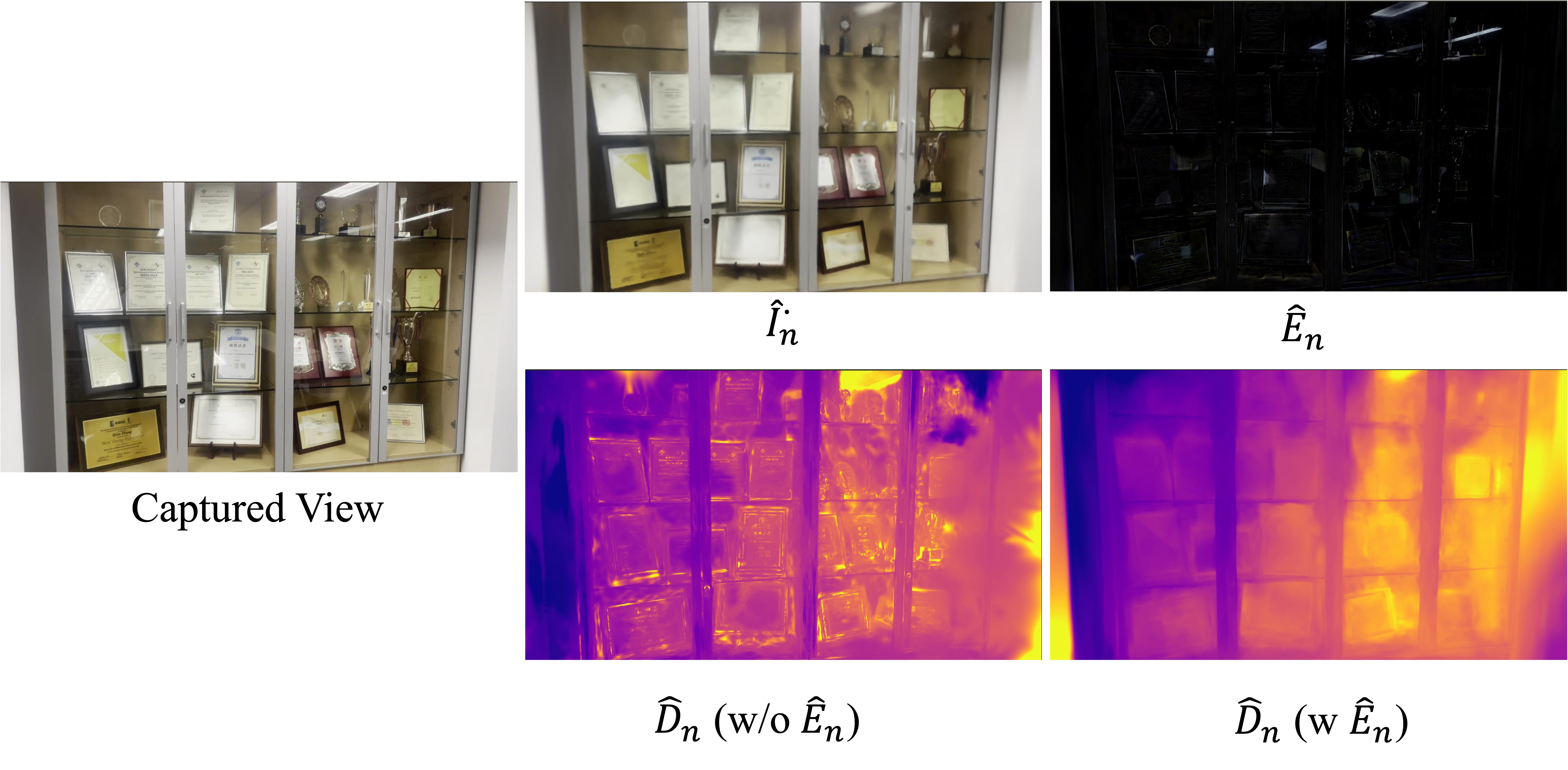}
    \vspace{-5mm}
    \caption{The significance of learnable residual maps during geometric prior utilization. Introducing residual maps significantly enhances the utilization of the geometric prior, resulting in cleaner, reflection-free depth maps.}
    \label{fig: residual_maps}
  \end{minipage}
  \hfill
  \begin{minipage}{0.45\textwidth}
    \centering
    \includegraphics[width=\textwidth]{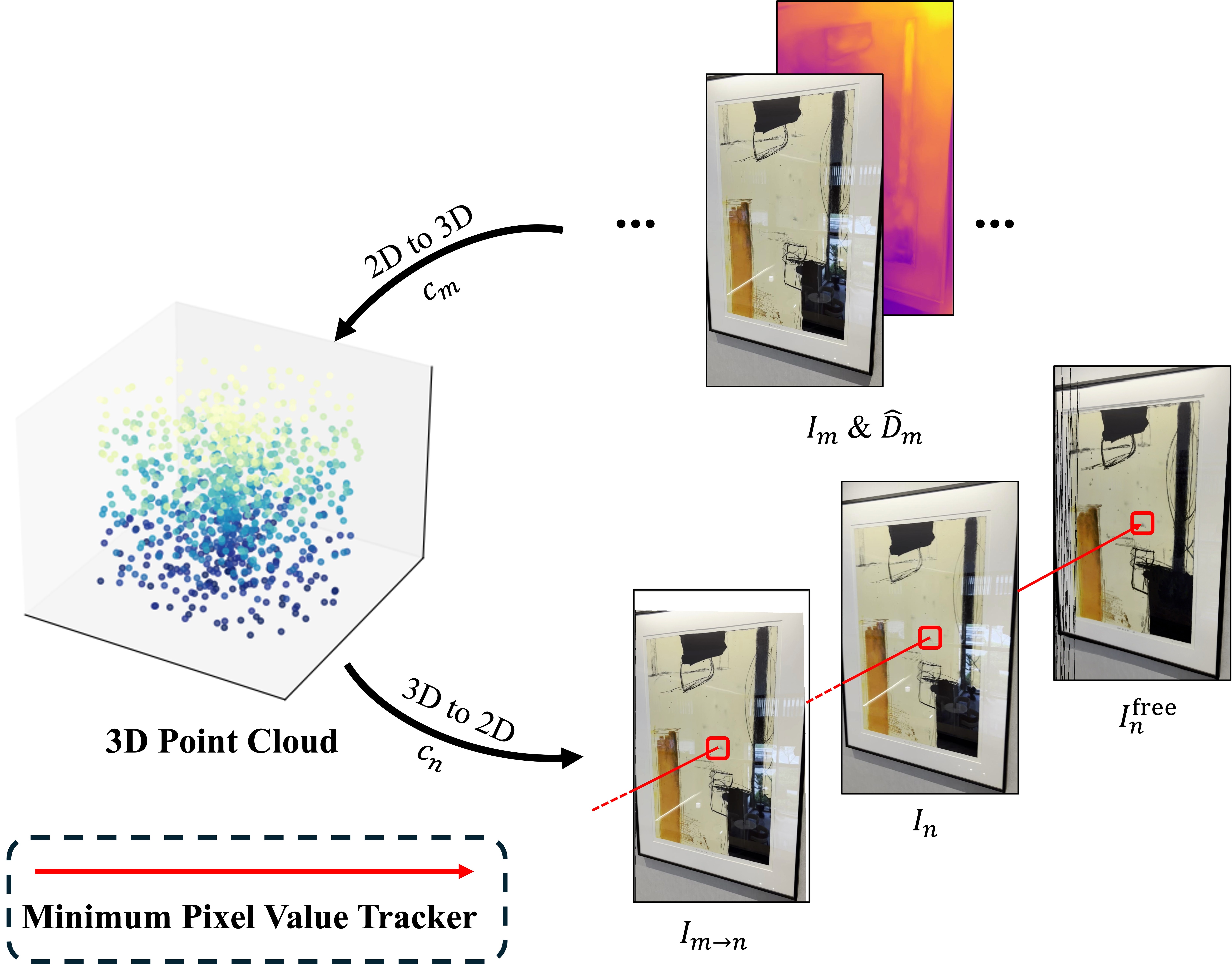}
    \caption{Illustration of depth-based 3D warp. $c_{*}$ denotes the camera parameters of the captured view $I_{*}$. For certain meaningless areas in the warped image $I_{m \to n}$, we fill these areas with pure white color.}
    \label{fig: 3D_warp}
  \end{minipage}
\end{figure}

Although vanilla 3DGS achieves impressive rendering quality for 3D reconstruction, it often produces floating artifacts and residual reflections in the final outputs during reconstructing a reflection-degraded scenario.
To mitigate these issues, we incorporate a geometric prior based on depth smoothness constraints~\cite{Guo_2022_CVPR}, which helps generate cleaner and more accurate depth maps.
Specifically, we apply a robust depth smoothness loss together with learnable residual maps that capture unwanted reflection components (see Fig.~\ref{fig: residual_maps}), enhancing the optimization of 3DGS on our reflection-degraded inputs:

\begin{equation}
\mathcal{L}_{\text{Geom}} = \mathcal{L}_{\text{vanilla}}(\hat{I}_{n}^{\cdot} + \tilde{E}_{n}, I_{n}) + \mathcal{L}_{\text{smooth}}(\hat{D}_{n}),
\end{equation}
where $\hat{E}{n}$ is the learned residual map, $\hat{D}{n}$ is the rendered depth map, and $\mathcal{L}_{\text{smooth}}$ is a bilateral filter loss~\cite{flash_splat} promoting local smoothness.

The key purpose of utilizing this geometric prior is to produce reflection-free depth maps with high accuracy and purity.
These improved depth maps are essential for the precise application of the photometric prior, allowing more reliable projection of pixel values into 3D space and better performance of the following photometric prior utilization.

\subsubsection{Photometric Prior Utilization}
Inspired by Li~\etal~\cite{li2013exploiting}, we observe that, for a point in a 3D scene, the minimum value across all captured views is highly correlated with representing reflection-free transmission.
Based on this photometric prior, for each reference view, we obtain its corresponding reflection-free version by the following operations: 1) Perform 3D warping of all auxiliary views using their corresponding over-smooth but reflection-free depth maps rendered from the geometric prior utilization; 2) Calculate the minimum value of all these warped auxiliary views and the reference view at each spatial position; 3) We treat this minimum map as the reflection-free version of the reference view, and regard it as our additional supervised signal during the following comprehensive optimization, a detailed illustration is shown in Fig.~\ref{fig: 3D_warp}.
These operations can be formulated as
\begin{equation}
I_{n}^{\text{free}} = \text{minimum}(I_{n}, \{\mathcal{W}(I_{m}, \hat{D}_{m}, c_{m \to n}) \}),
\end{equation}
where $m \in N$ and $m \neq n$, $\mathcal{W}(\cdot)$ denotes depth-based 3D warping function, $c_{*}$ denotes the camera parameters of the captured view $I_{*}$.
Although there are noticeable warping errors in $\mathcal{I}_{N}^{\text{free}}$ shown in Fig.~\ref{fig: 3D_warp}, we find that their reflection-free information contributes positively to the following optimization in 3DGS.
With a proposed ATR loss, we can effectively mitigate the impact of these errors, leading to improved final clean transmission results.

\subsubsection{Comprehensive Optimization}
To comprehensively optimize the 3DGS model for the final clean 3D scene, we initialize three 3DGS models for transmission, reflection and reflective fraction, respectively.
During optimization, we simultaneously optimize the three 3DGS models using several loss functions to achieve a balance, ensuring pure transmission and reflection with complete separation.

Firstly, the vanilla 3DGS loss based on Eq.~\ref{eq: degradation model} is applied:
\begin{equation}
\mathcal{L}_{\text{vanilla}} = \lambda_{\mathcal{L}_{1}} \mathcal{L}_{1}(\hat{I}_{n}^{\cdot\cdot}, I_{n}) + (1-\lambda_{\mathcal{L}_{1}}) \mathcal{L}_{\text{D-SSIM}}(\hat{I}_{n}^{\cdot\cdot}, I_{n}), 
\end{equation}
where $\hat{I}_{n}^{\cdot\cdot} = \hat{T}_{n} + \hat{\beta}_{n} \circ \hat{R}_{n}$, $\hat{T}_{n}$, $\hat{\beta}_{n}$ and $\hat{R}_{n}$ are rendered from their corresponding 3DGS.

Secondly, to give explicit guidance for optimizing the transmission 3DGS model, a naive L1 or MSE function can be applied to its rendered results: $\mathcal{L}_{\text{T}} = \mathcal{L}_{\mathcal{L}_{1} / \text{MSE}}(T_{n}, I_{n}) + \mathcal{L}_{\mathcal{L}_{1} / \text{MSE}}(T_{n}, I_{n}^{\text{free}})$.
However, during optimization, we find that the rendered result tends to be collapsed into one of $I_{n}$ (is fine-grained but suffers from reflection degradation) and $I_{n}^{\text{free}}$ (is free from reflections but exhibits artifacts).
Instead of empirically assigning reasonable weights to these two losses, we propose a more intuitive loss, \textbf{Adaptive Twin Rendering} (ATR) loss, that automatically balances them, enabling the transmission 3DGS model to effectively learn the optimal information from both sources:
\begin{equation}
\mathcal{L}_{\text{ATR}} = \text{minimum}(\left\| T_{n}-I_{n} \right\|_{2}^{2}, \left\| T_{n}-I_{n}^{\text{free}} \right\|_{2}^{2}).
\end{equation}
This loss works based on two facts: 1) Compared to reconstructing transmission and reflection simultaneously, a single 3DGS model tends to only reconstruct one of them at beginning, $\mathcal{L}_{\text{ATR}}$ encourages the 3DGS model to prioritize fitting the pure transmission information in $I_{n}^{\text{free}}$ rather than the more complex reflection-degraded information in $I_{n}$; 2) The 3DGS model tends to reconstruct 3D consistent information, the 3D inconsistent errors in $I_{n}^{\text{free}}$ are always ignored when there is more reasonable information in $I_{n}$.

Thirdly, to encourage the reflection 3DGS model to refrain from reconstructing transmission information, we adopt the exclusion loss~\cite{zhang2018single}, $\Phi_{\text{excl}}(I_{1}, I_{2})$, minimizing the similarity between the rendered transmission and reflection images.
This loss aims to ensure that the rendered outputs remain as distinct and pure as possible:
\begin{equation}
\mathcal{L}_{\text{excl}} = \mathcal{L}_{\text{excl}}(\hat{T}_{n}, \hat{R}_{n}) + \mathcal{L}_{\text{excl}}(I_{n}^{\text{free}}, \hat{R}_{n}).
\end{equation}

Fourthly, the bilateral filter loss for encouraging local smoothness is slightly imposed on rendered depths of transmission and reflection scenes, as well as reflective fraction maps. This loss is represented as $\mathcal{L}_{\text{smooth}}$.

Finally, we combine the above losses with corresponding weights for our comprehensive optimization:
\begin{equation}
\mathcal{L} = \lambda_{\text{1}}\mathcal{L}_{\text{vanilla}} + \lambda_{\text{2}}\mathcal{L}_{\text{ATR}} + \lambda_{\text{3}}\mathcal{L}_{\text{excl}} + \lambda_{\text{4}}\mathcal{L}_{\text{smooth}},
\end{equation}
where $\lambda_{\text{1}}=1.0$, $\lambda_{\text{2}}=1.0$, $\lambda_{\text{3}}=0.01$ and $\lambda_{\text{4}}=0.01$.

\section{3D Reflection Removal Benchmark}
\begin{table}[h]
    \centering
    \caption{Proposed 3D Reflection Removal Benchmark. 
    \textbf{Transparency}: Higher values indicate clearer transmission and weaker reflection. 
    \textbf{Thickness}: Affects light refraction and glass curvature. 
    \textbf{Reflection Strength}: Controls the intensity of the glass's reflection.}
    \vspace{-3mm}
    \label{tab:benchmark}
    \resizebox{0.4\textwidth}{!}{
    \begin{tabular}{ccc}
        \toprule
        \textbf{Configuration} & \textbf{Intensity Range} & \textbf{Total Cases} \\ 
        \midrule
        Transparency (\%)      & $[88, 92]$ & \multirow{3}{*}{45}              \\
        Thickness (mm)         & $[1, 7]$ &               \\
        Reflection Strength    & $[12, 135]$ &            \\
        \bottomrule
    \end{tabular}
    }
\end{table}
Although a real-world benchmark exists~\cite{zhu2022neural}, it comprises only four scenes, each featuring a single setting (e.g., glass transparency, thickness), which limits its dversity for comprehensive quantitative evaluation.
Besides, building a real-world 3D reflection removal benchmark with clean ground truth is challenging.
Capture multiple views, but each view is captured with and without reflection before the camera movement.
This necessitates multiple repositionings of the glass, resulting in spatial misalignment of the reflections.
This misalignment adversely impacts the quality of 3D reflection reconstruction, ultimately degrading the performance of the final removal.

Based on the abovementioned observation and inspired by Niklaus~\etal~\cite{niklaus2021learned}, we turn to build 3D reflection removal benchmark by the sophisticated rendering engine, Unity~\cite{haas2014history}.
With the high-quality reflection simulation in Unity, we find that the performance of 3D reflection removal methods on our proposed benchmark is highly consistent in real-world scenarios.
Precise camera movement control in Unity, combined with flexible reflection on/off switching, allows us to generate high-quality ground-truth transmission images for each view.
Considering the complexity of real-world scenarios, we build our benchmark considering various real-world reflecting factors, including the transparency, thickness and reflection strength of glasses provided in Unity, as shown in Table.~\ref{tab:benchmark}.
Ultimately, we construct 5 distinct 3D scenes by Unity, each with 9 variants, resulting in 45 cases in total for our proposed benchmark.

% \begin{table}[h]
%     \centering
%     \caption{Proposed 3D Reflection Removal Benchmark. 
%     \textbf{Transparency}: Higher values indicate clearer transmission and weaker reflection. 
%     \textbf{Thickness}: Affects light refraction and glass curvature. 
%     \textbf{Reflection Strength}: Controls the intensity of the glass's reflection.}
%     \vspace{-4mm}
%     \label{tab:benchmark}
%     \resizebox{0.45\textwidth}{!}{
%     \begin{tabular}{cccc}
%         \toprule
%         \textbf{Scenes} & \textbf{Configuration} & \textbf{Intensity Range} & \textbf{Total Cases} \\ 
%         \midrule
%         \multirow{3}{*}{5} & Transparency (\%)      & $[88, 92]$ & \multirow{3}{*}{45}              \\
%                           & Thickness (mm)         & $[1, 7]$ &               \\
%                           & Reflection Strength    & $[12, 135]$ &            \\
%         \bottomrule
%     \end{tabular}
%     }
% \end{table}

\begin{figure*}[t]
\includegraphics[width=\textwidth]{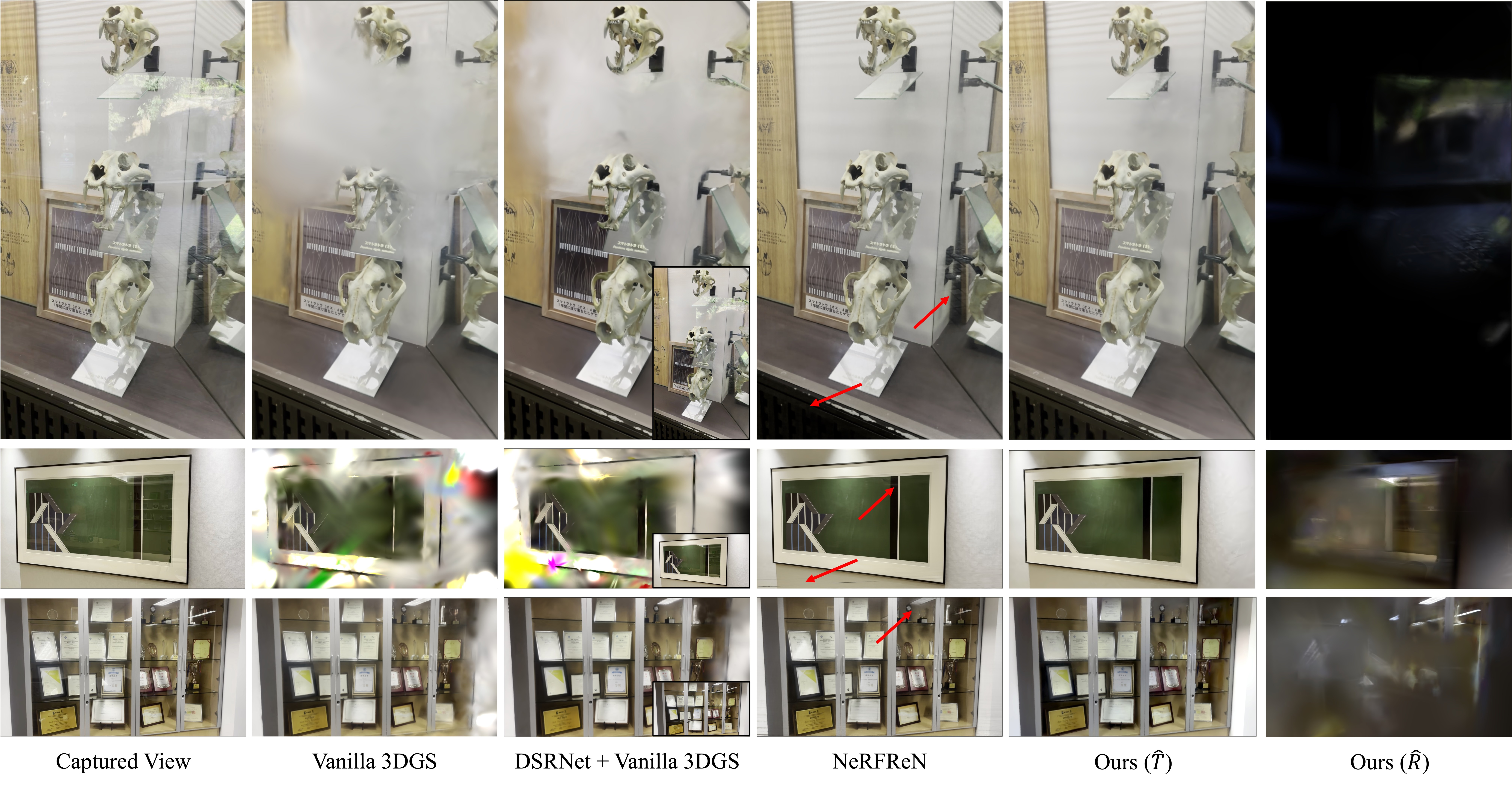}
\vspace{-8mm}
\caption{Qualitative evaluation on various real-world scenarios. ``DSRNet + Vanilla 3DGS'' indicates that degraded views are pre-processed using a state-of-the-art single-image method, DSRNet, before 3DGS optimization. For column 3, we show the view processed by DSRNet at the bottom right corner of each rendered result.}
\vspace{-5mm}
\label{fig: Real_comparison}
\end{figure*}

\section{Experiemnts}
\subsection{Implementation Details}
During the initialization, we employ VGGSfM~\cite{wang2024vggsfm} to create the initial point cloud. For a fair comparison, we consistently utilize VGGSfM across all 3D methods in this study.
During geometric prior utilization, the 3DGS model is optimized using the standard 3DGS settings as described by Kerbl~\etal~\cite{kerbl3Dgaussians}, with 5,000 iterations.
During comprehensive optimization, three separate 3DGS models are simultaneously optimized using the same hyperparameters but with different loss functions, each undergoing 30,000 iterations.

\subsection{Baselines}
We compare the proposed method against several strong baselines:
\begin{itemize}
\item 3D-based method: MVSNeRF, TNeRF, NeRFReN, vanilla 3DGS;
\item Burst-imaging-based method: Chugunov~\etal~\cite{chugunov2024neural};
\item Learning-based image method combined with 3DGS: DSRNet~\cite{hu2023single} + vanilla 3DGS.
\end{itemize}

\subsection{Quantitative Evaluation}
\begin{table}[t]
    \centering
    \vspace{-3mm}
    \caption{Quantitative evaluation on our proposed benchmark. ``DSRNet + Vanilla 3DGS'' indicates that degraded views are pre-processed using a state-of-the-art single-image method before 3DGS optimization. The burst-imaging-based method proposed by Chugunov~\etal~\cite{chugunov2024neural} fails to generate outputs with reasonable evaluation scores, indicating its inability to address the 3D reflection removal problem.}
    \vspace{-3mm}
    \label{tab:unity_metrics}
    \resizebox{0.45\textwidth}{!}{
    \begin{tabular}{c|ccc}
        \toprule
        \textbf{Method} & \textbf{PSNR} $\uparrow$ & \textbf{SSIM} $\uparrow$ & \textbf{LPIPS} $\downarrow$ \\
        \midrule
        Chugunov~\etal~\cite{chugunov2024neural} & 2.94 & 0.248 & 0.935 \\
        Vanilla 3DGS~\cite{kerbl3Dgaussians} & 29.58 & \textbf{0.917} & 0.175 \\
        DSRNet~\cite{hu2023single} + Vanilla 3DGS & 26.91 & 0.872 & 0.240 \\
        NeRFReN~\cite{Guo_2022_CVPR} & 26.36 & 0.755 & 0.284 \\
        Ours & \textbf{31.45} & 0.915 & \textbf{0.150} \\
        \bottomrule
    \end{tabular}
    }
\end{table}

\begin{table}[t]
    \centering
    \caption{Quantitative evaluation on the real-world benchmark~\cite{zhu2022neural}. Flash-Splat cannot work on such sRGB inputs, since it requires camera RAW images.}
    \vspace{-3mm}
    \label{tab:real_metrics}
    \resizebox{0.45\textwidth}{!}{
    \begin{tabular}{c|ccc}
        \toprule
        \textbf{Method} & \textbf{PSNR} $\uparrow$ & \textbf{SSIM} $\uparrow$ & \textbf{LPIPS} $\downarrow$ \\
        \midrule
        MVSNeRF~\cite{chen2021mvsnerf} & 17.83 & 0.685 & 0.400 \\
        NeRFReN~\cite{Guo_2022_CVPR} & 18.63 & 0.716 & 0.449 \\
        TNeRF~\cite{zhu2022neural} & 22.75 & 0.841 & 0.205 \\
        Vanilla 3DGS~\cite{kerbl3Dgaussians} & 21.73 & 0.835 & 0.238 \\
        DSRNet~\cite{hu2023single} + Vanilla 3DGS & 22.03 & 0.838 & 0.221 \\
        Flash-Splat~\cite{flash_splat} & N/A & N/A & N/A \\
        Ours & \textbf{25.04} & \textbf{0.915} & \textbf{0.190} \\
        \bottomrule
    \end{tabular}
    }
\end{table}

We conduct a comprehensive quantitative evaluation on both our proposed benchmark and the real-world benchmark~\cite{zhu2022neural} of various reflection removal methods, as presented in Table~\ref{tab:unity_metrics} and Table~\ref{tab:real_metrics}.

In Table~\ref{tab:unity_metrics}, the most notable observation is that the burst-imaging-based method proposed by Chugunov~\etal~\cite{chugunov2024neural} fails to generate outputs with reasonable evaluation scores, indicating its inability to address the 3D reflection removal problem. We also observe that despite being the state-of-the-art 3D reflection removal method, NeRFReN completely fails on ``Scene 3'', highlighting its lack of robustness across diverse scenarios. According to PSNR, which is highly sensitive to reflection degradation, the proposed method demonstrates strong superiority in 3D reconstruction with effective reflection removal. Furthermore, LPIPS scores indicate that the proposed method achieves state-of-the-art perceptual quality. However, due to SSIM's insensitivity to reflection degradation, the vanilla 3DGS achieves the highest scores. This suggests that the reflection removal priors employed by NeRFReN and the proposed method may slightly affect structural consistency.

In Table~\ref{tab:real_metrics}, the proposed method demonstrates significantly superior performance compared to previous works.
Notably, Flash-Splat fails to operate when only sRGB inputs are provided, as it requires camera RAW images.
This highlights that the RAW image requirement can be a burden for daily use.

\subsection{Qualitative Evaluation}
\begin{figure*}[t]
\includegraphics[width=\textwidth]{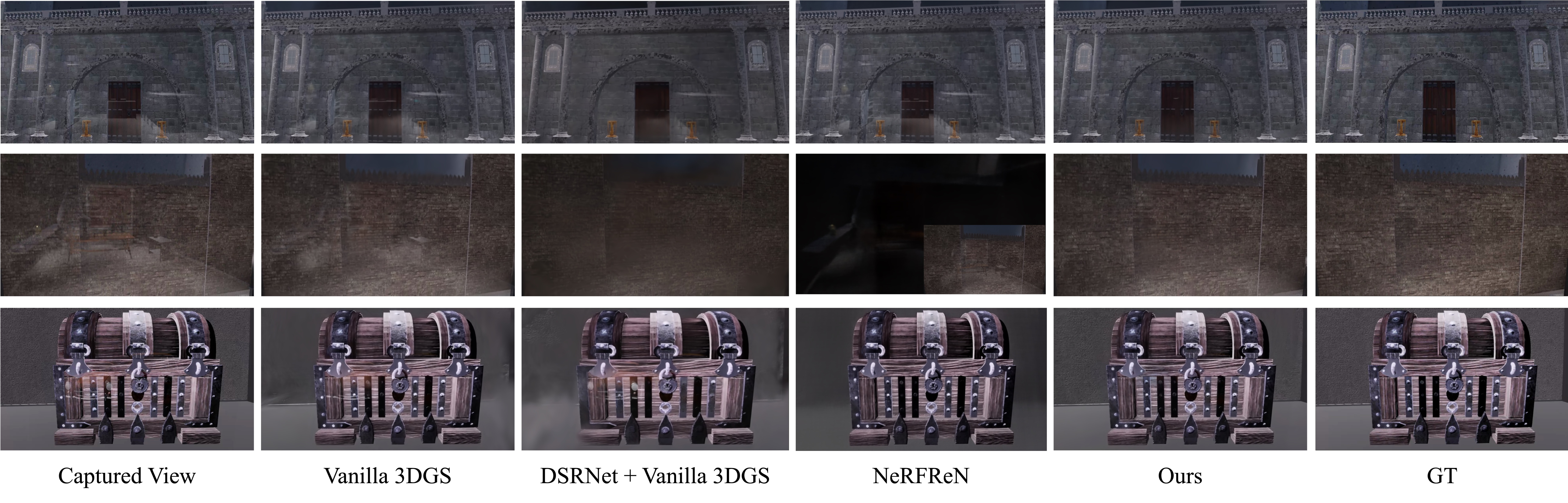}
\vspace{-5mm}
\caption{Qualitative evaluation on the proposed benchmark. ``DSRNet + Vanilla 3DGS'' indicates that degraded views are pre-processed using a state-of-the-art single-image method before 3DGS optimization. For row 2, as NeRFReN incorrectly identifies the transmission and reflection information, we included its predicted ``reflection layer'' in the bottom-right corner of the result.}
\label{fig: Unity_comparison}
\end{figure*}

\begin{figure*}[!h]
\includegraphics[width=\textwidth]{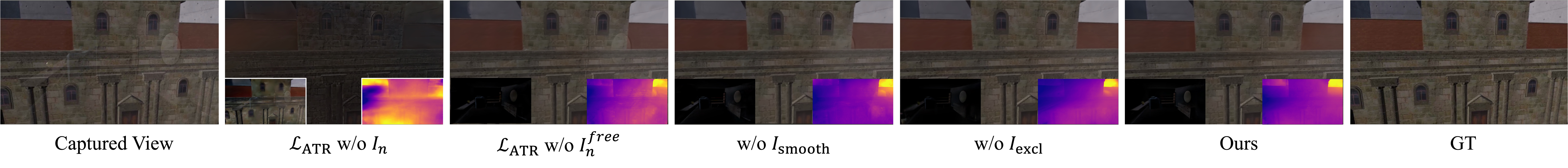}
\vspace{-5mm}
\caption{Qualitative evaluation of the ablation study. The reconstructed reflection scene and rendered depth map displayed in the small corner images.}
\label{fig: ablation}
\end{figure*}

In the qualitative evaluation, the proposed method showcases its strong superiority on our proposed benchmark. Furthermore, we also collect 6 real-world scenes for the qualitative evaluation.

In Fig.~\ref{fig: Real_comparison}, the proposed method demonstrates robust 3D reflection removal performance in real-world scenarios. As shown in column 2, the vanilla 3DGS approach struggles with severe floating artifacts during 3D reconstruction, heavily impacted by strong reflections. To address this, we applied a state-of-the-art image-based reflection removal method, DSRNet, prior to training the vanilla 3DGS. However, the limited generalization capability of the learning-based DSRNet results in suboptimal reflection removal for each captured view, ultimately leading to degraded 3D reconstruction performance, as illustrated in column 3. We observe that the 3D method, NeRFReN, struggles to remove strong reflections across all scenes, particularly with objects like lamps, where the proposed method excels. This clearly demonstrates the effectiveness of the photometric prior utilization in the proposed method. Additionally, due to the pixel-wise supervision in NeRFReN, the method lacks satisfactory global perceptual quality in the rendered results, leading to inconsistent pixel values across all outputs. Furthermore, the proposed method also reconstructs the underlying 3D reflection scenes, which is a potential guidance for the downstream task.

In Fig.~\ref{fig: Unity_comparison}, column 3, we observe that while the single-image method, DSRNet, significantly enhances the reflection removal performance for vanilla 3DGS, it also compromises the clarity of the transmission details, particularly the high-frequency textures on the walls and objects. Due to the domain gap issue, such learning-based methods lack sufficient robustness across diverse scenes, resulting in residual reflections in the final outputs.
In column 4, NeRFReN fails to effectively remove reflections while preserving clear transmission details. Additionally, it faces challenges in distinguishing between transmission and reflection information, particularly in the scene depicted in row 2.
The proposed method effectively removes reflections while preserving high-frequency textures in the transmission information, a particularly challenging aspect of the reflection removal task.

\subsection{Ablation Study}
We conduct a comprehensive ablation study to assess the impact of each key component in our proposed method, as shown in Table~\ref{tab: ablation} and Fig.~\ref{fig: ablation}. Firstly, we observe that the ``$\mathcal{L}_{\text{ATR}}$ w/o $I_{n}$'' variant, which relies solely on $I_{n}^{\text{free}}$ as the supervised signal, suffers from severe floating artifacts and color distortion due to the photometric prior, leading to unsatisfactory evaluation scores. Next, we analyze the ``$\mathcal{L}_{\text{ATR}}$ w/o $I_{n}^{\text{free}}$'' variant, which undergoes full optimization but uses only $I_{n}$ as the supervised signal, lacking supervision from the $I_{n}^{\text{free}}$. Although this variant achieves acceptable 3D reconstruction performance, its reflection removal capability is significantly compromised. These findings underscore the effectiveness of the proposed ATR loss, demonstrating that relying solely on $I_{n}$ or $I_{n}^{\text{free}}$ as the supervised signal fails to achieve a satisfactory balance between 3D reconstruction and reflection removal. Finally, we evaluate the effectiveness of the exclusion loss and smooth loss using two distinct variants, as shown in Table~\ref{tab: ablation} and Fig.~\ref{fig: ablation}. Both losses contribute to improved overall performance. The ``w/o $\mathcal{L}_{\text{smooth}}$'' variant demonstrates that without the local smoothness constraint on depth maps, strong reflections are more likely to persist in the rendered transmission image. The ``w/o $\mathcal{L}_{\text{excl}}$'' variant reveals that, without the exclusion loss, the reflection scene may retain transmission information, resulting in incomplete transmission reconstruction, as shown in the bottom-left corner of the rendered result.
We also investigate the effect of the number of input views on the proposed method and observe that it maintains satisfactory robustness across different input view counts.

\begin{table}[!h]
    \centering
    \caption{Quantitative evaluation of the ablation study.}
    \vspace{-3mm}
    \label{tab: ablation}
    \resizebox{.4\textwidth}{!}{
    \begin{tabular}{c|ccc}
        \toprule
        Method & PSNR $\uparrow$ & SSIM $\uparrow$ & LPIPS $\downarrow$\\
        \midrule % 
        \multicolumn{1}{l|}{Ours ($\mathcal{L}_{\text{ATR}}$ w/o $I_{n}$)} & 19.64 & 0.790 & 0.237 \\
        \multicolumn{1}{l|}{Ours ($\mathcal{L}_{\text{ATR}}$ w/o $I_{n}^{\text{free}}$)} & 30.45 & 0.888 & 0.176 \\
        \multicolumn{1}{l|}{Ours (w/o $\mathcal{L}_{\text{excl}}$)} & 31.00 & 0.908 & 0.155 \\
        \multicolumn{1}{l|}{Ours (w/o $\mathcal{L}_{\text{smooth}}$)} & 31.11 & 0.909 & 0.155 \\
        \multicolumn{1}{l|}{Ours} & \textbf{31.25} & \textbf{0.915} & \textbf{0.150} \\
        \midrule % 
        \multicolumn{1}{l|}{Ours (11 views)} & 31.14 & 0.907 & \textbf{0.140} \\
        \multicolumn{1}{l|}{Ours (22 views)} & \textbf{31.27} & 0.912 & 0.146 \\
        \multicolumn{1}{l|}{Ours (88 views)} & 31.25 & \textbf{0.915} & 0.150 \\
        \bottomrule
    \end{tabular}
    }
\end{table}

\subsection{Processing Time} 
Table~\ref{tab:time} presents a detailed comparison of training times on the proposed benchmark with 88 views per case. Our method requires 60.74 minutes to train, which is significantly faster than NeRFReN’s 63.48 hours, demonstrating a substantial improvement in efficiency. While Vanilla 3DGS and DSRNet combined with Vanilla 3DGS complete training more quickly (5.94 and 7.85 minutes respectively), these methods do not achieve the same level of performance and robustness as our method. Overall, our method strikes a strong balance between training speed and quality, offering a practical solution for large-scale or time-sensitive applications.

\begin{table}[!h]
    \centering
    {\setlength{\tabcolsep}{0.03cm} % local scope for tabcolsep
    \setlength{\arrayrulewidth}{1.0pt}
    \caption{The processing-time comparison on the proposed benchmark. \textbf{Time} represents the training time for each case (88 views).\label{tab:time}}
    \vspace{-3mm}
    \resizebox{.45\textwidth}{!}{
    \begin{tabular}{
        >{\centering\arraybackslash}m{1cm} |
        >{\centering\arraybackslash}m{2cm}
        >{\centering\arraybackslash}m{3cm}
        >{\centering\arraybackslash}m{1.4cm}
        >{\centering\arraybackslash}m{1.6cm}
        >{\centering\arraybackslash}m{1.4cm}
        }
        \toprule
        & Vanilla 3DGS
        & \makecell{DSRNet + \\ Vanilla 3DGS}
        & NeRFReN
        & Ours
        & \makecell{Ours \\ (3D Warp)} \\
        \cmidrule(lr){1-6}
        Time & 5.94 min & 7.85 min & 63.48 h & 60.74 min & 30 sec \\
        \bottomrule
    \end{tabular}}
    }
\end{table}

\section{Conclusion}
In this work, we introduce a novel reflection-free 3DGS method for the 3D reflection removal task, named RF3DGS. Unlike previous works that rely on additional photography techniques, such as flash or no-flash cues, RAW images, the proposed method simply requires a monocular sRGB video capturing multiple views of a 3D scene. By leveraging the photometric prior, we propose an adaptive twin rendering (ATR) loss that effectively enhances both reflection removal and 3D reconstruction performance. Additionally, we present a 3D benchmark with substantial diversity to enable quantitative evaluation of 3D reflection removal methods. Extensive experiments on the proposed benchmark and real-world scenes demonstrate the strong superiority of the proposed method over existing methods.

{
    \small
    \bibliographystyle{ieeenat_fullname}
    \bibliography{main}
}

% Supplementary intentionally omitted for the arXiv main body; see original project if needed.
% \input{sec/X_suppl}

\end{document}